%% file: bmvc_final.tex
\documentclass{bmvc2k}
\usepackage{times}
\usepackage{epsfig}
\usepackage{graphicx}
\usepackage{amsmath}
\usepackage{amssymb}
\usepackage{lipsum}
\usepackage{caption} 
\usepackage{multirow}
\usepackage{booktabs}
\usepackage{pifont}
\usepackage{soul}
\usepackage{enumitem}
\newcommand{\xmark}{\ding{55}}%

\title{AISFormer: Amodal Instance Segmentation with Transformer}

\addauthor{Minh Tran}{minht@uark.edu}{1}
\addauthor{Khoa Vo}{khoavoho@uark.edu}{1}
\addauthor{Kashu Yamazaki}{kyamazak@uark.edu}{1}
\addauthor{Arthur Fernandes}{Arthur.Fernandes@cobbvantress.com}{2}
\addauthor{Michael Kidd}{mkidd@uark.edu}{3}
\addauthor{Ngan Le}{thile@uark.edu}{1}

\addinstitution{
 Department of CSCE, University of Arkansas\\
}

\addinstitution{
 Cobb-Vantress, Inc
}
\addinstitution{
Poultry Science, University of Arkansas\\
}
\runninghead{Minh Tran Et al.}{AISFormer}

\begin{document}

\maketitle
\begin{abstract}

Amodal Instance Segmentation (AIS) aims to segment the region of both visible and possible occluded parts of an object instance. While Mask R-CNN-based AIS approaches have shown promising results, they are unable to model high-level features coherence due to the limited receptive field. The most recent transformer-based models show impressive performance on vision tasks, even better than Convolution Neural Networks (CNN). In this work, we present AISFormer, an AIS framework, with a Transformer-based mask head. AISFormer explicitly models the complex coherence between occluder, visible, amodal, and invisible masks within an object's regions of interest by treating them as learnable queries. 
Specifically, AISFormer contains four modules: (i) feature encoding: extract ROI and learn both short-range and long-range visual features. (ii) mask transformer decoding: generate the occluder, visible, and amodal mask query embeddings by a transformer decoder (iii) invisible mask embedding:  model the coherence between the amodal and visible masks, and (iv) mask predicting: estimate output masks including occluder, visible, amodal and invisible. We conduct extensive experiments and ablation studies on three challenging benchmarks i.e. KINS, D2SA, and COCOA-cls to evaluate the effectiveness of AISFormer. The code is available at: \url{https://github.com/UARK-AICV/AISFormer}

\end{abstract}


\section{Introduction}

\begin{figure*}[t]
    \centering
    \includegraphics[width=0.95\textwidth]{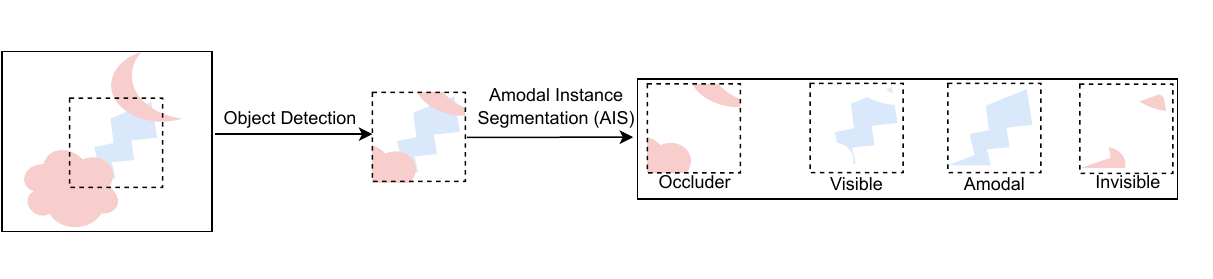}
    \vspace{0.05cm}
    \caption{An explanation of different mask instances in Amodal Instance Segmentation (AIS). Given a region of interest (ROI) extracted by an object detector, AIS aims to extract both visible and invisible mask instances including occluder, visible, amodal, and invisible.} 
    \label{fig:AIS}
    \vspace{-0.5cm}
\end{figure*}

Instance segmentation (IS) is a fundamental task in computer vision, which predicts each object instance and its corresponding per-pixel segmentation mask. Modern IS approaches \cite{lee2020centermask, xie2020polarmask, ghiasi2021simple, zhang2021refinemask} are typically built on CNN and follow Mask R-CNN \cite{he2017mask} paradigm with two stages of first detecting bounding boxes, and then segmenting instance masks. Mask R-CNN has been improved by various backbone architecture designs to obtain high accuracy IS performance. Despite demonstrating impressive performance, those Mask R-CNN-based approaches are limited when dealing with occlusion, in which a lot of segmentation errors happen due to overlapping objects, especially among object instances belonging to the same class \cite{ke2021deep}. The capability of perceiving/segmenting the entirety of an object instance regardless of partial occlusion is known as amodal instance perception/segmentation (AIS). In AIS, there are various mask instances taken into consideration including occluder, visible, amodal, and invisible as shown in Fig.\ref{fig:AIS}. 

\begin{figure*}[t]
    \centering
    \includegraphics[width=1.\textwidth]{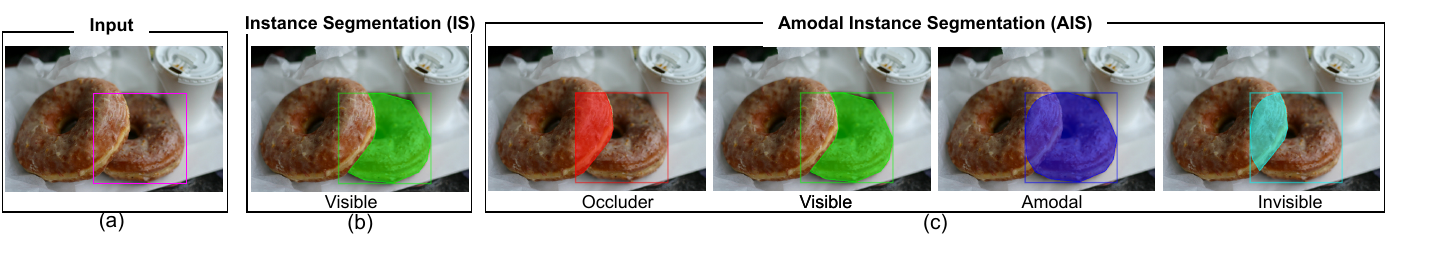}
    \caption{A comparison between Instance Segmentation (IS) and Amodal Instance Segmentation (AIS). Given an image with ROI (a), IS aims to extract the visible mask instance (b) whereas AIS aims to extract both the visible mask and occluded parts (c).} 
    \label{fig:IS-AIS}
    \vspace{-0.5cm}
\end{figure*}

A comparison between IS and AIS is given in Fig.\ref{fig:IS-AIS}. While IS aims to segment the visible region of the donut, as shown in the middle panel, AIS aims to extract both visible and occluded regions of the donut as shown in the right panel of Fig.\ref{fig:IS-AIS}. Humans are capable of amodal perception, but computers are limited to modal perception \cite{valada2016convoluted}, which has played an important role in theoretical study and real-world applications. In the literature, various AIS approaches \cite{follmann2019learning, qi2019amodal, li2016amodal, xiao2021amodal, ke2021deep} have been based on the IS pipeline of Mask R-CNN \cite{he2017mask} and attempted to predict the amodal mask of objects given the pooling region of interest (ROI) feature. In such networks, the main contribution comes from segmentation head designs, with attention paid to the amodal instance mask regression after obtaining the ROI. 
For example, \cite{follmann2019learning} performs AIS through an amodal branch and invisible mask loss, \cite{qi2019amodal} reasons invisible parts via a multi-task framework with Multi-Level Coding (MLC), \cite{xiao2020amodal} proposes a cross-task attention based refinement strategy to refine the amodal mask and the visible mask, \cite{ke2021deep} introduces a bilayer decoupling layer extract to occluder and infer occludee. 
The Mask R-CNN-based AIS approaches have shown promising AIS results, 
however, they are unable to model high-level or long-range features coherence due to the limited receptive field. That means the complex relationship between ROIs has been not modeled in such Mask R-CNN-based AIS approaches. 

Recently, the DETR\cite{carion2020end}-based approaches \cite{dong2021solq, fang2021instances, hu2021istr} have shown their advantage and achieved on-par performances compared with CNN-based ones on object detection and instance segmentation. Indeed, they model the long-range relationships between objects across visual features by reformulating objects as learnable query embeddings and take advantage of the transformer attention mechanism.
Inspired by that advantage, we propose AISFormer, an AIS framework, with a Transformer-based mask head. Our AISFormer consists of four modules: feature encoding, mask transformer decoder, invisible mask embedding, and segmentation. In the first module, after obtaining the region of interest (ROI) feature from the backbone and ROIAlign algorithm, CNN-based layers and a transformer encoder are applied to learn both short-range and long-range features of the given ROI. The second module aims to generate the occluder, visible and amodal mask query embeddings by a transformer decoder. The next module extracts invisible mask embeddings to model the coherence between the amodal mask and visible mask. The last module, the segmentation module, estimates the output masks including occluder, visible, amodal, and invisible.
Our overall network is shown in Fig.\ref{fig:network}. Our contribution can be summarized as follows:
\begin{itemize}[noitemsep,topsep=0pt]
    \item We propose AISFormer, an amodal instance segmentation framework, with a Transformer-based mask head. Our AISFormer can explicitly model the complex coherence between occluder, visible, amodal, and invisible masks within an object's regions of interest by treating them as learnable queries. AISFormer also models the relationship between these embeddings and the corresponding region of interest.
    \item We empirically validate the usefulness of our proposed method by showing that it achieves superior performance to most of the current state-of-the-art methods benchmarked on  three amodal datasets, i.e., KINS \cite{qi2019amodal}, COCOA-cls \cite{zhu2017semantic}, and D2SA \cite{follmann2018mvtec}.
\end{itemize}
\vspace{-0.3cm}
\section{Related Work}
\vspace{-0.3cm}
\subsection{Instance Segmentation (IS)} The existing IS approaches can be summarized into two categories of two-stage and one-stage. The former category, two-stage IS approaches, is divided into two groups: top-down and bottom-up methods based on the priority of detection and segmentation. The top-down two-stage approaches  \cite{cai2018cascade, chen2019tensormask, chen2019hybrid, cheng2020boundary} follow the pipeline of Mask R-CNN \cite{he2017mask} by first detecting bounding boxes and then performing segmentation in each RoI region. The bottom-up two-stage approaches \cite{arnab2016bottom, chen2017deeplab, newell2017associative} first generate semantic segmentation and then group pixels through clustering or metric learning techniques. While the two-stage category has not taken the correlation between detection and segmentation into consideration, the latter category, single-stage, addresses this limitation with less time consumption \cite{bolya2019yolact}. There are two common groups in the single-stage category i.e., anchor-based and anchor-free methods. The anchor-based one-stage approaches \cite{li2017fully, chen2019tensormask, bolya2019yolact} simultaneously generate a set of class-agnostic candidate masks on the candidate region and extract instances from a semantic branch. To cope with the drawback of anchor-based approaches that rely heavily on pre-defined anchors sensitive to hyper-parameters, anchor-free one-stage approaches \cite{ying2019embedmask, chen2020blendmask,lee2020centermask} are proposed by eliminating anchor boxes and using corner/center points.
\vspace{-0.3cm}
\subsection{Amodal Instance Segmentation (AIS)}
Unlike IS, which only focuses on visible regions, AIS predicts both the visible and amodal masks of each object instance. One of the first works in AIS is proposed by Li~\textit{et~al.}~\cite{li2016amodal}, which relies on the directions of high heatmap values computed for each object to iteratively enlarge the corresponding object modal bounding box. Generally, AIS can be categorized into two groups of weakly supervised and fully supervised methods.

\noindent
\textbf{Fully supervised AIS}: The majority of AIS approaches follow a fully supervised approach, where the amodal annotation is required. The amodal annotation can come from human labeling \cite{follmann2019learning, qi2019amodal, xiao2020amodal}  or synthetic occlusions \cite{zhan2020self}. 
Follmann~\textit{et~al.}~\cite{follmann2019learning} propose ORCNN, in which they have two main instances mask heads, amodal and inmodal. In addition, they build another mask head on top of these two for invisible mask prediction. Based on ORCNN architecture, Qi~\textit{et~al.}~\cite{qi2019amodal} introduce ASN, where they propose the multi-level coding module. The module aims to learn global information, thus obtaining better segmentation for the invisible part. Different from ORCNN and ASN, VQ-VAE~\cite{jang2020learning} replaces the fully convolutional instance mask heads with variational autoencoders. Specifically, the input feature is firstly classified into an intermediate shape code. Then, the shape code is reconstructed into a complete object shape. Xiao~\textit{et~al.}~\cite{xiao2021amodal} also uses a pre-trained autoencoder to refine the initial mask prediction from the amodal mask head. The pre-trained autoencoder encodes the mask to a shape prior memory codebook and decodes it into a refined amodal mask. More recently, BCNet\cite{ke2021deep} attempts to decouple the mixing features by extending another branch that predicts the mask of occluders inside the bounding box of the occludee. Thus, the model can be aware of occluders and more precisely predict the amodal mask.\\
\noindent
\textbf{Weakly supervised AIS} (WAIS): \cite{zhan2020self, nguyen2021weakly} consider this problem under weak supervision where they attempt to generate amodal mask ground-truths from datasets using the available visible ground truths. The reason is that
human annotators sometimes could not provide reliable ground truth on amodal masks. The procedure of \cite{zhan2020self, nguyen2021weakly} on WAIS can be described as follows. First, data augmentation is applied to generate occlusion on top of objects in the dataset. Then, a UNet model \cite{ronneberger2015u} is trained as a segmenter. The input of the UNet is the available visible mask ground truth of the object. However, with the occlusion augmentation added to the object, the visible ground truth is treated as the amodal ground truth. In the second training stage, the amodal segmentations outputs from the trained UNet in the first phase are taken as a pseudo-ground truth for learning a standard instance segmenter – Mask R-CNN \cite{ke2022mask}. In the inference phase, Mask R-CNN trained on the generated amodal ground truth is expected to yield the correct AIS.  Another approach in WAIS uses object bounding boxes as inputs \cite{sun2020weakly, sun2022amodal}, where shape priors and probabilistic models are applied to classify visible, invisible, and background parts within the bounding box.\\
\noindent
Our ASIFormer belongs to the first group of fully supervised learning, in which visible, amodal, and occluder masks are represented by queries in a unified manner. 
\begin{figure*}[t]
    \centering
    \includegraphics[width=1.\textwidth]{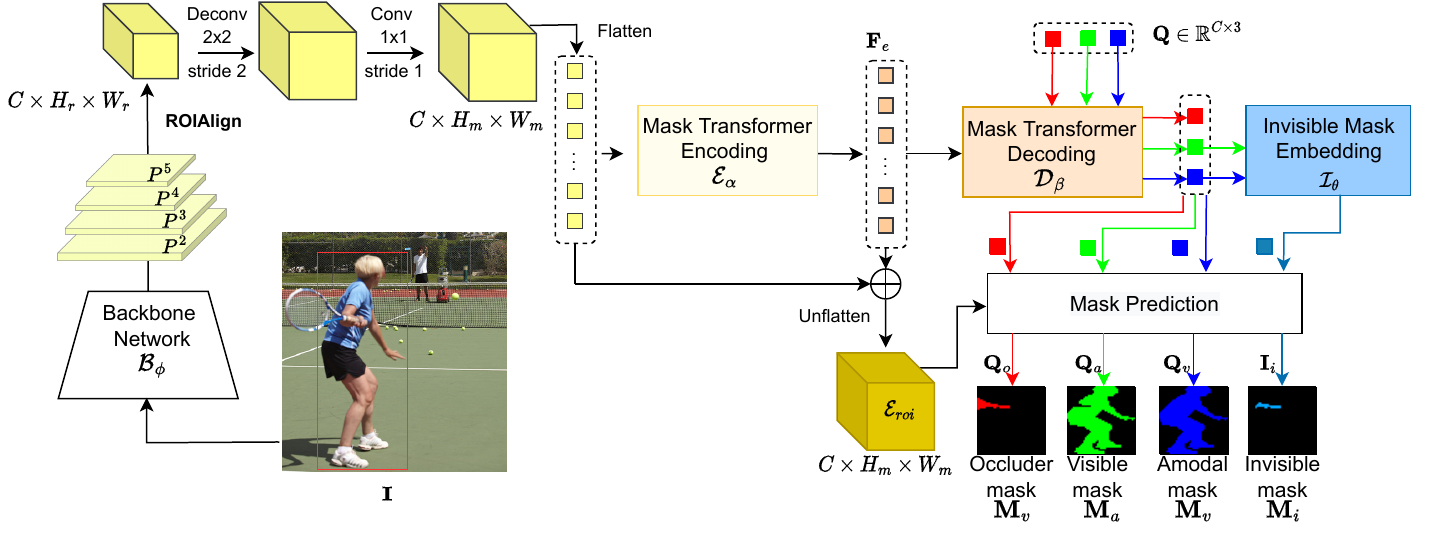}
    \caption{The overall flowchart of our proposed AISFormer. AISFormer consists of four modules corresponding to (i) feature encoding: after obtaining the region of interest (ROI) feature from the backbone $\mathcal{B}_\phi$ and ROIAlign algorithm $\varphi$, CNN-based layers and a transformer encoder are applied to learn both short-range and long-range features of the given ROI.(ii) mask transformer decoding $\mathcal{D}_\beta$: generate the occluder, visible, and amodal mask query embeddings by a transformer decoder (iii) invisible mask embedding $\mathcal{I}_\theta$ to model the coherence between the amodal and visible masks by computing the invisible mask embedding, and (iv) segmentation to estimate output masks including occluder ($\mathbf{M}_o$), visible ($\mathbf{M}_v$), amodal ($\mathbf{M}_a$)and invisible ($\mathbf{M}_i$).} 
    \label{fig:network}
    \vspace{-0.3cm}
\end{figure*}
\vspace{-0.3cm}
\subsection{Query-based Image Segmentation}
Inspired by DETR \cite{carion2020end}, query-based IS methods \cite{dong2021solq, fang2021instances, hu2021istr} have achieved impressive IS performance by treating segmentation as a set prediction problem. 
In these methods, queries are used to represent the objects and jointly learn multiple tasks of classification, detection, and segmentation.
In \cite{dong2021solq}, each query represents one object with multiple representations of class, location, and mask. For the IS branch, the mask encoding vectors are learnt by the compression coding (DCT \cite{ahmed1974discrete}, PCA \cite{abdi2010principal}, Sparse Coding \cite{donoho2006compressed}) of raw spatial masks. Instead of high-dimensional masks, ISTR  \cite{hu2021istr} predicts low-dimensional mask embeddings and is trained by a set loss based on bipartite matching.
Meanwhile, QueryInst \cite{fang2021instances} addresses IS problem by using parallel dynamic mask heads \cite{jia2016dynamic}. In QueryInst, queries are shared between tasks of detection and segmentation in each stage via dynamic convolutions. Unlike the previous works which process on regular dense tensors, Mask Tranfiner \cite{ke2022mask} first decomposes and represents an image region as a hierarchical quadtree. Then, all points on the quadtree are transformed into a query sequence to predict labels. By doing that, Mask Tranfiner only processes detected error-prone tree nodes, which are mostly along object boundaries or in high-frequency regions, and self-corrects their errors in parallel.
Similar to \cite{dong2021solq, fang2021instances, hu2021istr}, our proposed AISFormer is inspired by DETR. Indeed, our learnable queries represent visible, amodal, and occluder masks. Our transformer-based approach aims to model the coherence between them.

\vspace{-0.3cm}
\section{Methods}
We propose an approach, termed AISFormer, to efficiently tackle AIS. The overall architecture of AISFormer is illustrated in Fig \ref{fig:network}.
Our AISFormer consists of four modules corresponding to (i) feature encoding: in this first module, after obtaining the region of interest (ROI) feature from the backbone $\mathcal{B}_\phi$ and ROIAlign algorithm $\varphi$, CNN-based layers and a transformer encoder are applied to learn both short-range and long-range features of the given ROI.
(ii) mask transformer decoding $\mathcal{D}_\beta$: generate the occluder, visible, and amodal mask query embeddings by a transformer decoder (iii) invisible mask embedding $\mathcal{I}_\theta$ to model the coherence between the amodal and visible masks by computing the invisible mask embedding, and (iv) segmentation to estimate output masks including occluder, visible, amodal and invisible.

Generally, given an image $\mathbf{I}$, AIS aims to predict the amodal mask $\mathbf{M}_a$ and visible mask $\mathbf{M}_v$ within ROIs. To better model the mask coherence between amodal mask $\mathbf{M}_a$ and visible mask $\mathbf{M}_v$, our AISFormer estimates other relevant masks such as occluder mask $\mathbf{M}_o$ and invisible mask $\mathbf{M}_i$. In such relationships, the amodal mask $\mathbf{M}_a$ consists of the visible part $\mathbf{M}_v$ and the invisible mask $\mathbf{M}_i$, which is a part of occluder mask $\mathbf{M}_o$. 
\vspace{-0.3cm}
\subsection{Feature Encoding}
Given an image $\mathbf{I}$, we first adopt a pre-trained image recognition network $\mathcal{B}_\phi$ (e.g. ResNet \cite{he2017mask}, RegNet\cite{radosavovic2020designing}) to extract the spatial visual features, where $\phi$ is backbone network weights. After backbone feature extraction, we apply ROIAlign algorithm $\varphi$ to obtain ROIs corresponding to the bounding boxes in the image $\mathbf{F}_{roi} = \varphi(\mathcal{B}_\phi(\mathbf{I}))$, where $\mathbf{F}_{roi} \in \mathbb{R}^{C\times H_r\times W_r}$. To represent $\mathbf{F}_{roi}$ in  higher resolution without increasing computational cost, we apply a deconvolution layer with a stride of 2 into the spatial domain. As a result, the feature maps are upsampled, i.e., $\mathbf{F}_{roi} \in \mathbb{R}^{C\times H_m\times W_m}$, whereas $H_m = 2 \times H_r$ and $W_m = 2 \times W_r$. The higher-resolution feature maps $\mathbf{F}_{roi}$ is passed through a  $1\times1$ stride $1$ convolutional layer and then flattened from the spatial dimensions into one dimension 
and obtain $H_rW_r$ embeddings with the dimension of $C$. The  flattened feature maps with their positional encodings \cite{vaswani2017attention} are fed into the mask transformer encoder $\mathcal{E}_\alpha$ to extract long-range dependency across spatial domains, where $\alpha$ is transformer encoder network weights. The mask transformer encoder is designed as a self-attention model and described in Fig.\ref{fig:design}(a). The output of this module is denoted as $\mathbf{F}_e$ and computed as:
\begin{equation}
\mathbf{F}_e = \mathcal{E}_\alpha(\text{Flatten}(\mathbf{F}_{roi})) \text{; where } \mathbf{F}_{roi} = \varphi(\mathcal{B}_\phi(\mathbf{I}))
\end{equation}
\vspace{-0.9cm}
\subsection{Mask transformer decoding}
Inspired by DETR\cite{carion2020end}, our mask transformer decoder transforms $3$ learnable mask query-embeddings $\mathbf{Q}_o, \mathbf{Q}_v, \mathbf{Q}_a$, corresponding to occluder, visible and amodal masks, respectively.
This mask transformer decoder $\mathcal{D}_\beta$ is designed as in the Fig.\ref{fig:design} (b) which combines both self-attention ($\mathcal{A}_{self}$) and cross-attention($\mathcal{A}_{cross}$). The self-attention takes three learnable mask query-embeddings $\mathbf{Q} = [\mathbf{Q}_o, \mathbf{Q}_v, \mathbf{Q}_a] \in \mathbb{R}^{C \times 3}$ as input to model the correlation between queries. Meanwhile, the cross attention takes the output from the self-attention model $\mathcal{A}_{self}$, encoding feature $\mathbf{F}_e$ (i.e. the output of the mask transformer encoder) and their positional encodings as input to learn the coherence between individual query and feature maps $\mathbf{F}_e$. The mask transformer decoding $\mathcal{D}_\beta$ is described as follows.
\begin{equation}
    \mathbf{Q} = \mathcal{D}_\beta(\mathbf{Q}, \mathbf{F}_e) =  \mathcal{A}_{cross}\left(\mathbf{F}_e,( \mathcal{A}_{self}(\mathbf{Q}) \oplus  \mathbf{Q})\right)
\end{equation}
\vspace{-0.9cm}
\subsection{Invisible mask embedding}
\label{sec:imp}
\begin{figure*}[t]
    \centering
    \includegraphics[width=.95\textwidth]{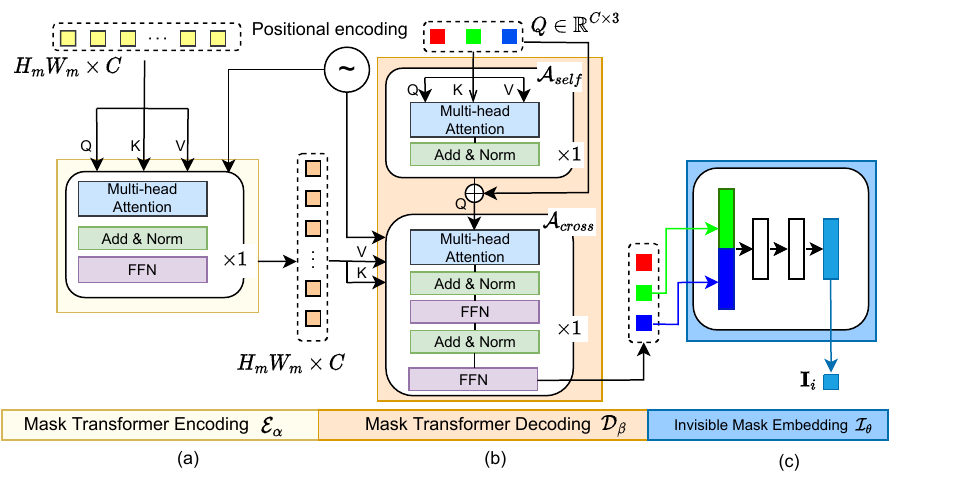}
    \caption{Illustration network architecture of AISFormer. (a): mask transformer encoder $\mathcal{E}_\alpha$ is designed as one block of self-attention, (b): mask transformer decoder $\mathcal{D}_\beta$ is designed as a combination of one block of self-attention ($\mathcal{A}_{self}$) and one block of cross-attention ($\mathcal{A}_{cross}$) and (c): invisible embedding  $\mathcal{I}_\theta$ is designed as an MLP with two hidden layers. } 
    \label{fig:design}
    \vspace{-0.5cm}
\end{figure*}

Our observation is that there are correlations between the visible mask and amodal mask as given in Fig.\ref{fig:AIS}. Specifically, the amodal mask is considered as the union of the visible mask (i.e modal mask) and the invisible mask (i.e. occluded parts). In order to model that relationship, we propose an invisible mask embedding module $\mathcal{I}_\theta$, which takes visible and amodal mask query instances (i.e. $\mathbf{Q}_v, \mathbf{Q}_a$) from the previous module $\mathcal{D}_\beta$ into consideration. The mask instances $\mathbf{Q}_v, \mathbf{Q}_a$) are concatenated into a long vector in $\mathbb{R}^{2C}$. The invisible mask embedding module $\mathcal{I}_\theta$ is designed as in Fig.\ref{fig:design}(c) with an MLP consisting of hidden layers. The output $\mathbf{I}_i \in \mathbb{R}^C$ is computed as follows:
\begin{equation}
    \mathbf{I}_i = MLP([\mathbf{Q}_v; \mathbf{Q}_a])
\end{equation}
\vspace{-0.9cm}
\vspace{-0.3cm}
\subsection{Mask predicting}
The mask prediction module aims to predict four relevant mask instances as defined in AIS, i.e. occluder ($\mathbf{M}_o$), visible ($\mathbf{M}_v$), amodal ($\mathbf{M}_a$), and invisible ($\mathbf{M}_i$) masks. In this module, we first compute per-pixel ROI embeddings $\mathcal{E}_{roi} \in \mathbb{R}^{C\times H_m \times W_m}$. Indeed, $\mathcal{E}_{roi}$ represents the features of each pixel in the set of output masks $\mathbf{M} = [\mathbf{M}_o, \mathbf{M}_v, \mathbf{M}_a, \mathbf{M}_i]$. We define the per-pixel ROI embeddings as in the following equation.
\begin{equation}
    \mathcal{E}_{roi} = \text{Unflatten}\left(\text{Flatten}(\mathbf{F}_{roi}) \oplus  \mathbf{F}_e\right)
\end{equation}
We then apply a dot product ($\otimes$) on mask query embeddings $\mathbf{Q} = [\mathbf{Q}_o, \mathbf{Q}_v, \mathbf{Q}_a]$, invisible mask embedding $\mathbf{I}_i$ and per-pixel ROI embeddings $\mathcal{E}_{roi}$ to obtain the output masks $\mathbf{M}$ as in Eq.\ref{eq:M}. Our AISFormer is designed as an end-to-end framework and trained using the cross-entropy loss.

\begin{equation}
    [\mathbf{M}_o, \mathbf{M}_v, \mathbf{M}_a, \mathbf{M}_i] = \mathcal{E}_{roi} \otimes [\mathbf{Q}_o, \mathbf{Q}_v, \mathbf{Q}_a, \mathbf{I}_i]
\label{eq:M}
\end{equation}
\vspace{-0.9cm}
\section{Experiments}
\subsection{Datasets, Metrics and Implementation Details}

\noindent
\textbf{Datasets:}
We benchmark our AISFormer and conduct the comparison with other SOTA methods on three amodal segmentation datasets, namely KINS \cite{qi2019amodal}, COCOA-cls \cite{zhu2017semantic}, and D2SA \cite{follmann2018mvtec}. KINS is a large-scale traffic dataset with modal and amodal annotation masks of instances and is created based on the KITTI dataset \cite{geiger2013vision}. Our AISFormer is trained on the training split with 7,474 images (95,311 instances) and tested on the testing split with 7,517 images (92,492 instances). 
D2SA is built based on the D2S(Densely Segmented Supermarket) \cite{follmann2018mvtec} dataset with 60 categories of instances related to supermarket items. There are 2,000 training images and 3,600 testing images. The annotations are generated by overlapping objects on others. COCOA-cls is created based on COCO dataset \cite{lin2014microsoft}. In COCOA-cls, there are 2,476 training images and 1,223 testing images. It also contains 80 instance classes about common objects in context.

\noindent
\textbf{Metrics:} In order to conduct the evaluation of AISFormer and comparison with other existing approaches, we adopt mean average precision (AP) and mean average recall (AR), commonly used evaluation metrics in AIS tasks.

\noindent
\textbf{Implementation Details:}
We implement our AISFormer based on Detectron2 \cite{wu2019detectron2}, a Pytorch-based detection framework. For the KINS dataset, we use an SGD optimizer with a learning rate of 0.0025 and a batch size of 1 on 48000 iterations. For D2SA datasets, we also train with an SGD optimizer but with a learning rate of 0.005 and a batch size of 2 on 70000 iterations. For COCOA-cls, since it has a smaller amount of training images, we train it on 10000 iterations with the learning rate of 0.0005 and a batch size of 2. The experiments have been conducted using an Intel(R) Core(TM) i9-10980XE 3.00GHz CPU  and a Quadro RTX 8000 GPU.

\vspace{-0.3cm}
\subsection{Performance Comparison}
Table \ref{tab:kins} compares our AISFormer with SOTA AIS methods on the KINS dataset. AISFormer obtains steady improvement on distinct backbones (i.e. ResNet-50, ResNet-101, and RegNet). In particular, in comparison with the existing methods evaluated on KINS using ResNet-50 as the backbones, our method outperforms both SOTA methods with shape prior (e.g., VQVAE \cite{jang2020learning} by $3.5\% AP $ and VRSP-Net \cite{xiao2021amodal} by $1.7\% AP$ ) and without using shape priors (e.g., Mask R-CNN \cite{he2017mask} by $3.8\% AP$ and ASN \cite{qi2019amodal} by $1.6\% AP$), respectively. In addition, our method outperforms previous methods using ResNet-101 as the backbone by a large margin from $2\%-4.4\% AP$ and has a comparable result with the current SOTA APSNet \cite{mohan2022amodal}, using RegNet \cite{radosavovic2020designing} as the backbone.
\begin{table}
\caption{Performance comparison on KINS dataset. $\dagger$ indicates our reproduced results. On each backbone, the best scores are in \textbf{bold} and the second best scores are in \underline{underlines}.}
\vspace{0.15cm}
\centering\resizebox{\columnwidth}{!}{
\begin{tabular}{c|c|c|c|c|cccc}
\toprule
Sup. & Backbone & Model& Venue & \shortstack{Shape Prior}  & $AP$ & $AP_{50}$ & $AP_{75}$ & $ AR$
\\ \toprule
\multirow{2}{*}{\rotatebox{0}{\shortstack{Weakly}}}
& ResNet-50& PCNet~\cite{zhan2020self} & CVPR`20 & \xmark& 29.1 & 51.8 & 29.6 & 18.3 \\
& ResNet-50& ABSU~\cite{nguyen2021weakly} & ICCV`21 & \checkmark & 29.3 & 52.1 & 29.7 & 18.4 \\
\midrule
\multirow{11}{*}{\rotatebox{0}{Fully}}
&ResNet-50 &VQVAE~\cite{jang2020learning} & IEEE TMI`20& \checkmark & 30.3 & -- & -- & -- \\
&ResNet-50&VRSP-Net~\cite{xiao2021amodal} & AAAI`21 & \checkmark & \underline{32.1} & \underline{55.4} & \underline{ 33.3} & \underline{20.9}  \\ \cline{2-9}
& ResNet-50& Mask R-CNN~\cite{he2017mask} & ICCV`17 & \xmark& 30.0 & 54.5 & 30.1 & 19.4 \\
&ResNet-50&ORCNN~\cite{follmann2019learning}  & WACV`19& \xmark &  30.6 & 54.2 & 31.3 & 19.7 \\
&ResNet-50&ASN~\cite{qi2019amodal}& CVPR`19 &  \xmark & 32.2 & -- & -- & --\\
& ResNet-50&\textbf{AISFormer} & -- & \xmark &  \textbf{33.8}& \textbf{57.8} & \textbf{35.3} & \textbf{21.1}\\
\cmidrule{2-9}
& ResNet-101&Mask R-CNN~\cite{he2017mask}$^\dagger$ & ICCV`17&  \xmark&  30.2 & 54.3 & 30.4& 19.5 \\
& ResNet-101 &BCNet \cite{ke2021deep}  & CVPR`21 & \xmark &28.9& -- & -- & --\\
& ResNet-101 &BCNet \cite{ke2021deep} $^\dagger$ & CVPR`21 & \xmark & \underline{32.6} & \underline{57.2} & \underline{35.4} & \underline{21.5}\\
&ResNet-101&\textbf{AISFormer}  & --& \xmark &  \textbf{34.6} &\textbf{58.2}& \textbf{36.7} &\textbf{21.9} \\
\cmidrule{2-9}
& RegNet &APSNet~\cite{mohan2022amodal}  & CVPR`22 &  \xmark & \textbf{35.6} & -- & -- & -- \\
&RegNet &\textbf{AISFormer}  &--& \xmark & \shortstack{\textbf{35.6}}  & \textbf{59.9}&\textbf{37.0} &\textbf{22.5}  \\
\bottomrule
\end{tabular}
\label{tab:kins}}
\vspace{-0.3cm}
\end{table}

We also compare our model with all existing SOTA methods on the D2SA and COCOA-cls datasets, using ResNet-50 as the backbone as shown in Table \ref{tab:d2sa}. Compare with other existing shape-free approaches, our AISFormer achieves the SOTA performance on both datasets with a large margin. Specifically, compared with the second best i.e. BCNet \cite{ke2021deep}, our AISFormer gains $1.25\% AP$ and $1.23\% AR$ on D2SA and $0.63\% AP$ and $0.91\% AR$ on COCOA-cls. Notably, our AISFormer outperforms shape prior VRSP-Net \cite{xiao2021amodal} on the COCOA-cls dataset while obtaining high performance and comparable results with VRSP-Net \cite{xiao2021amodal} on the D2SA dataset. Our observation and hypothesis on the D2SA dataset are that the majority of objects in D2SA are supermarket items and their shapes are standardized and unvarying (e.g. items are in the shape of a rectangle or tube). Thus, with standardized shape objects, the shape prior approach i.e. VRSP-Net \cite{xiao2021amodal} outperforms other shape-free approaches; however, it does not work well on other datasets which contain more variety of shapes. Generally, AIS performance on the D2SA dataset is higher than KINS and COCOA-cls datasets, which contain more variety of objects and shapes.
\begin{table}
\caption{Performance comparison on the D2SA and COCOA-cls datasets with ResNet-50 as the backbone. $\dagger$ indicates our reproduced results. In the category of without shape prior, the best scores are in \textbf{bold} and the second best scores are in \underline{underlines}.}
\vspace{0.20cm}
\centering
\resizebox{\columnwidth}{!}{
\begin{tabular}{c|c|c|cccc|cccc}
\toprule
\multirow{2}{*}{Model} & \multirow{2}{*}{Venue} &  \multirow{2}{*}{Shape Prior} & \multicolumn{4}{c}{D2SA} & \multicolumn{4}{|c}{COCOA-cls}\\
\cline{4-11}
 & & &$AP$ & $AP_{50}$ & $AP_{75}$ & $ AR$ & $AP$ & $AP_{50}$ & $AP_{75}$ & $AR$\\
\midrule
VRSP-Net~\cite{xiao2021amodal}& AAAI`21 & \checkmark &{70.27}& {85.11}& {75.81}& {69.17}& {35.41} & 56.03 & {38.67} & {37.11} \\ \hline
Mask R-CNN~\cite{he2017mask} & ICCV`17 &\xmark&63.57& 83.85& 68.02& 65.18 & 28.03& 53.68& 25.36& 29.83 \\
ORCNN~\cite{follmann2019learning} &WACV`19 &\xmark & 64.22 & 83.55 &69.12& 65.25  & 28.03& 53.68& 25.36& 29.83 \\
ASN~\cite{qi2019amodal} $^\dagger$ &CVPR`19 & \xmark  &  63.94&\textbf{84.35}  &{69.57} & 65.20 &\underline{35.33} & \underline{58.82} &\underline{37.10} & 35.50 \\
BCNet \cite{ke2021deep} $^\dagger$ & CVPR`21 &\xmark  & \underline{65.97}  & \underline{84.23} &\underline{72.74} &\underline{66.90}& 35.14 & \textbf{58.84} & 36.65 &\underline{35.80} \\
\textbf{AISFormer} & -- & \xmark & \textbf{67.22} & {84.05} & \textbf{72.87} & \textbf{68.13} & \textbf{35.77}& {57.95}& \textbf{38.23}& \textbf{36.71}\\
\bottomrule
\end{tabular}
}
\label{tab:d2sa}
\end{table}
\vspace{-0.3cm}
\subsection{Ablations}
\label{text:ablations}
We further investigate the effectiveness of our proposed AISFormer. We conduct six experiments as shown in  Table \ref{tab:embeddings} as follows:

\noindent
$\bullet$ \textbf{Exp \#1}: AISFormer with only one amodal mask query embeddings (i.e. $\mathbf{Q} = [\mathbf{Q}_a$]).

\noindent
$\bullet$ \textbf{Exp \#2}: AISFormer with amodal mask query and occluder query embeddings (i.e. $\mathbf{Q} = [\mathbf{Q}_a, \mathbf{Q}_o$]). By adding occluder query embedding, the performance improves 2.46\% AP on D2SA and 1.9\% AP on KINS. This shows the effectiveness of occluder query embedding.

\noindent
$\bullet$ \textbf{Exp \#3}: AISFormer with amodal mask query and visible query embeddings (i.e. $\mathbf{Q} = [\mathbf{Q}_a, \mathbf{Q}_v$]). By adding visible query embedding, the performance improves 1.90\% AP on D2SA and 1.69\% AP on KINS. This shows the effectiveness of visible query embedding.

\noindent
$\bullet$ \textbf{Exp \#4}: AISFormer with amodal mask query, visible query embeddings (i.e. $\mathbf{Q} = [\mathbf{Q}_a, \mathbf{Q}_v, \mathbf{Q}_i$]), and invisible query embedding. This shows the effectiveness of invisible mask embedding, which helps to gain 0.97\% AP on D2SA and 0.23\% AP on KINS, in comparison with Exp \#3.

\noindent
$\bullet$ \textbf{Exp \#5}: Both amodal mask query, visible query and occluder embeddings (i.e. $\mathbf{Q} = [\mathbf{Q}_o, \mathbf{Q}_a, \mathbf{Q}_v$]) are used and it obtains better performance on each individual embedding. This shows the effectiveness of learnable query embeddings.  

\noindent
$\bullet$ \textbf{Exp \#6}: This is our proposed network architecture, AISFormer, which obtains the best performance. This shows the effectiveness of both learnable query embeddings and invisible mask embedding when combining them into an end-to-end framework.

We further visualize the attention maps of learnable queries as in Fig.\ref{fig:attn}. More qualitative results are included in the supplementary.


\begin{table}[!h]
	\caption{Effectiveness of occluder mask embeddings (Exps. \#1 vs. \#2, \#3 vs. \#5), visible mask embeddings (Exps. \#1 vs.  \#3,  \#2 vs. \#5), and invisible mask embeddings (Exps. \#3 v.s.\#4, \#5 v.s.\#6) with ResNet-50 as the backbone on the D2SA and KINS datasets.}
 \vspace{0.2cm}
	\centering
\resizebox{0.8\linewidth}{!}{
		\begin{tabular}{ c | c | c |c | c | c | c | c | c }
			\toprule
			\multirow{2}{*}{Exp.} & \multirow{2}{*}{Amodal} &  \multirow{2}{*}{Occluder} &
			\multirow{2}{*}{Visible} & \multirow{2}{*}{Invisible} & \multicolumn{2}{c|}{D2SA} & \multicolumn{2}{c}{KINS}\\
			\cline{6-9} 
			& & & & & ${AP}$ & ${AP}_{50}$&  ${AP}$ & ${AP}_{50}$   \\
			\midrule 
			\#1 & \checkmark & \xmark  & \xmark & \xmark& 63.55&80.23 & 31.01&54.28 \\
			\#2 & \checkmark & \checkmark  & \xmark & \xmark&66.01&82.89&32.91& 55.39\\
			\#3 & \checkmark & \xmark  & \checkmark &  \xmark&65.45&81.97 &32.70&55.16\\
			\#4 & \checkmark &\xmark  & \checkmark & \checkmark& 66.42&83.62&32.93&55.45\\
			\#5 & \checkmark & \checkmark&\checkmark &\xmark &67.00 &83.50 & 33.21&56.20 \\
			\#6 & \checkmark & \checkmark&\checkmark &\checkmark &67.22&84.05 &33.78 & 57.80 \\
			\bottomrule
		\end{tabular}
		}
	\label{tab:embeddings}
 \vspace{-0.5cm}
\end{table}
\begin{figure*}[t]
    \centering
    \includegraphics[width=1.\textwidth]{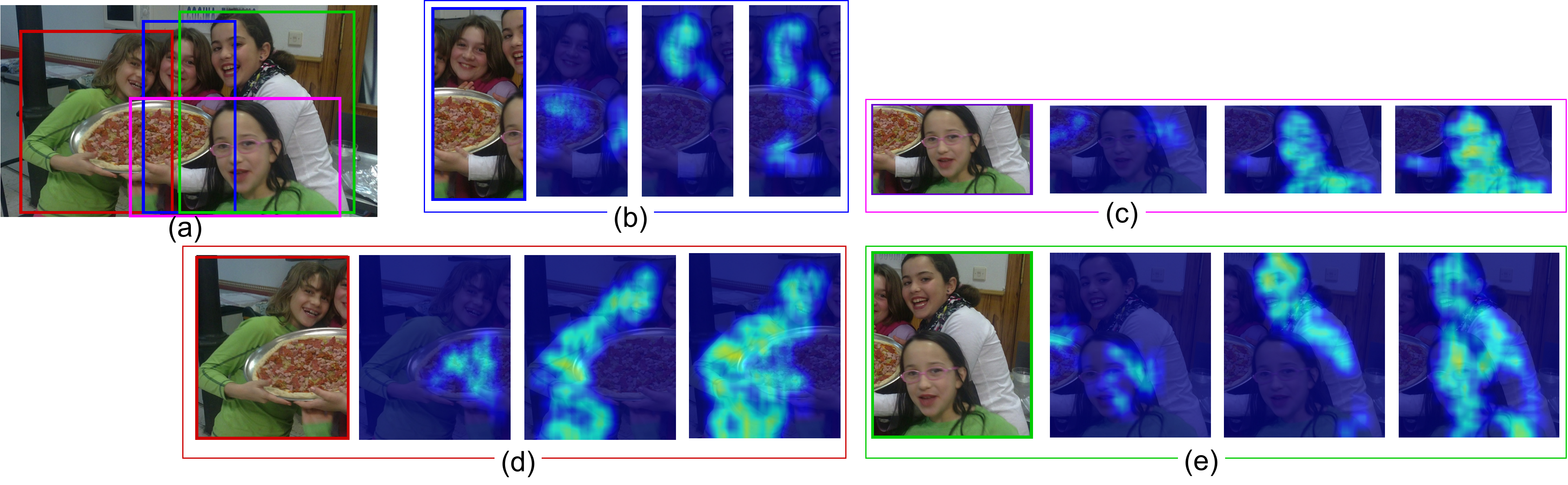}
    \vspace{0.2cm}
    \caption{Attention visualization of query embeddings. (a): Input image with four ROIs. (b), (c), (d), (e): attention feature maps of queries in each ROI. For each ROI, from left-right: ROI, occluder query embedding, visible query embedding, and amodal query embedding.} 
    \label{fig:attn}
    \vspace{-0.3cm}
\end{figure*}

\subsection*{Conclusion}
\vspace{-0.1cm}
We have proposed AISFormer, an AIS framework, with Transformer-based mask heads. The proposed is designed as an end-to-end framework with four modules: feature encoding, mask transformer decoding, invisible mask embedding, and mask predicting. Our AISFormer can explicitly model the complex coherence between occluder, visible, amodal, and invisible mask instances within ROI by treating them as learnable queries. Results and ablation study on three benchmarks, namely KINS, D2SA, and COCOA-cls, show the effectiveness of our proposed learnable queries and indicate that our AISFormer obtains the SOTA AIS performance compared to the existing methods. Our future research will integrate the shape prior into AISFormer as well as investigate AISFormer on other modalities such as time-lapse, and videos.

\subsection*{Acknowledgments}
This material is based upon work supported by the National Science Foundation (NSF) under Award No OIA-1946391, NSF 1920920, and NSF 2223793.

\bibliography{bmvc_final}

\clearpage
\input{tmp/bmvc_supp}

\end{document}

%% file: tmp/bmvc_supp.tex
\appendix
\renewcommand{\thetable}{\Roman{table}}
\renewcommand{\thefigure}{\Roman{figure}}
\setcounter{table}{0}
\setcounter{figure}{0}

\begin{center}{\bf \Large Appendix}\end{center}\vspace{-2mm}
In this supplementary, we provide more qualitative performance obtained by our proposed AISFormer. More specifically, we will illustrate the following:
\begin{itemize}
    \item Qualitative results conducted by AISFormer compared to Mask R-CNN \cite{he2017mask} and BCNet \cite{ke2021deep} on the KINS \cite{qi2019amodal}, D2SA \cite{follmann2018mvtec}, and COCOA-cls\cite{zhu2017semantic} datasets.
    \item Attention map visualizations indicating the cross attention between our learnable queries and the corresponding region of interest.
\end{itemize}


\subsection*{Qualitative results}
\begin{figure*}[t]
    \centering
    \includegraphics[width=\textwidth]{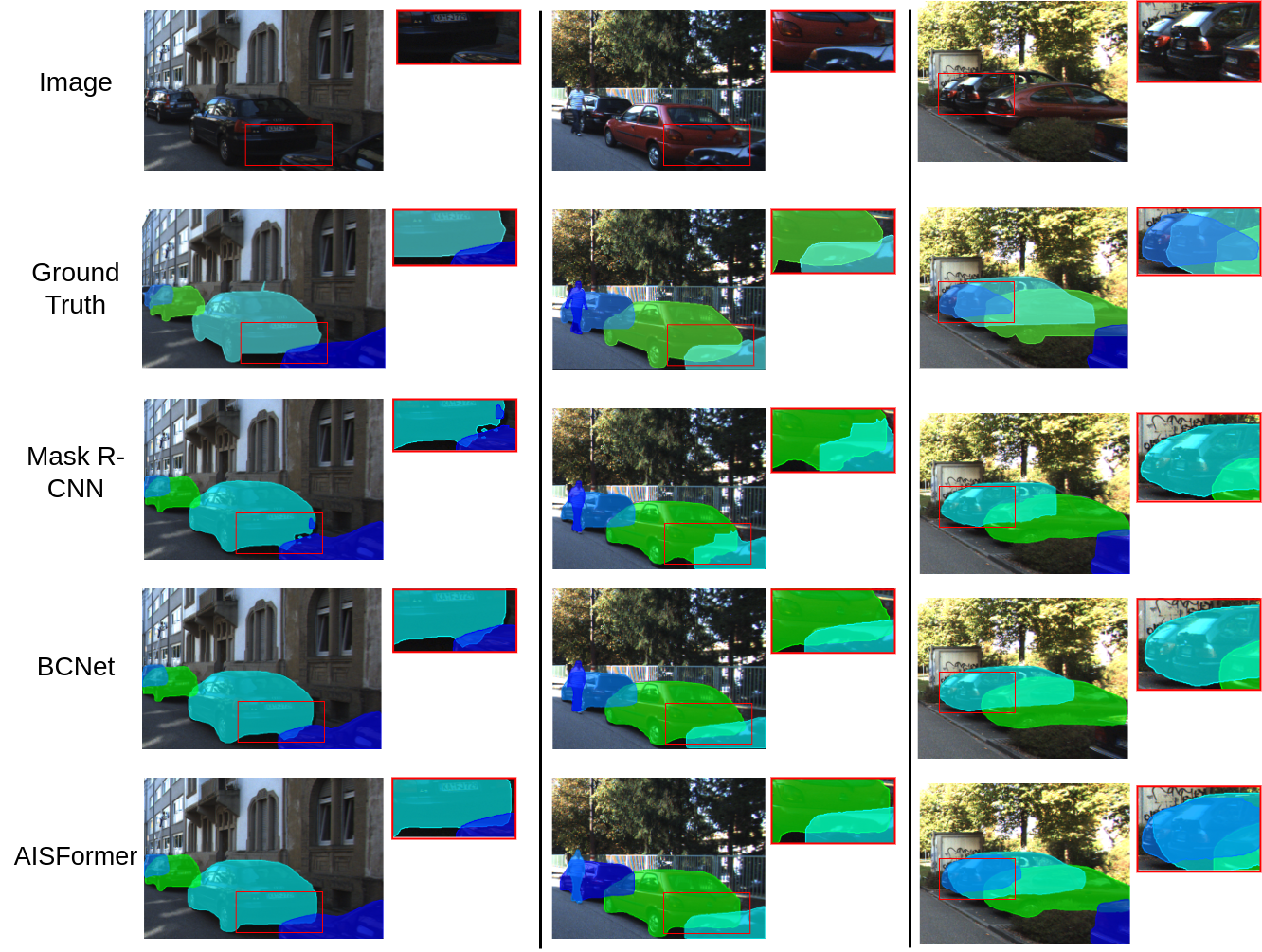}
    \vspace{0.2cm}
    \caption{Qualitative Results on KINS dataset. The rows from top to bottom are the images, the ground-truth amodal masks, and estimations of Mask R-CNN, BCNet, and our AISFormer, respectively. Differences between methods are best shown in red box regions. Best view in zoom and color.} 
    \label{fig:qual_kins}
    \vspace{-0.3cm}
\end{figure*}

\begin{figure*}[t]
    \centering
    \includegraphics[width=\textwidth]{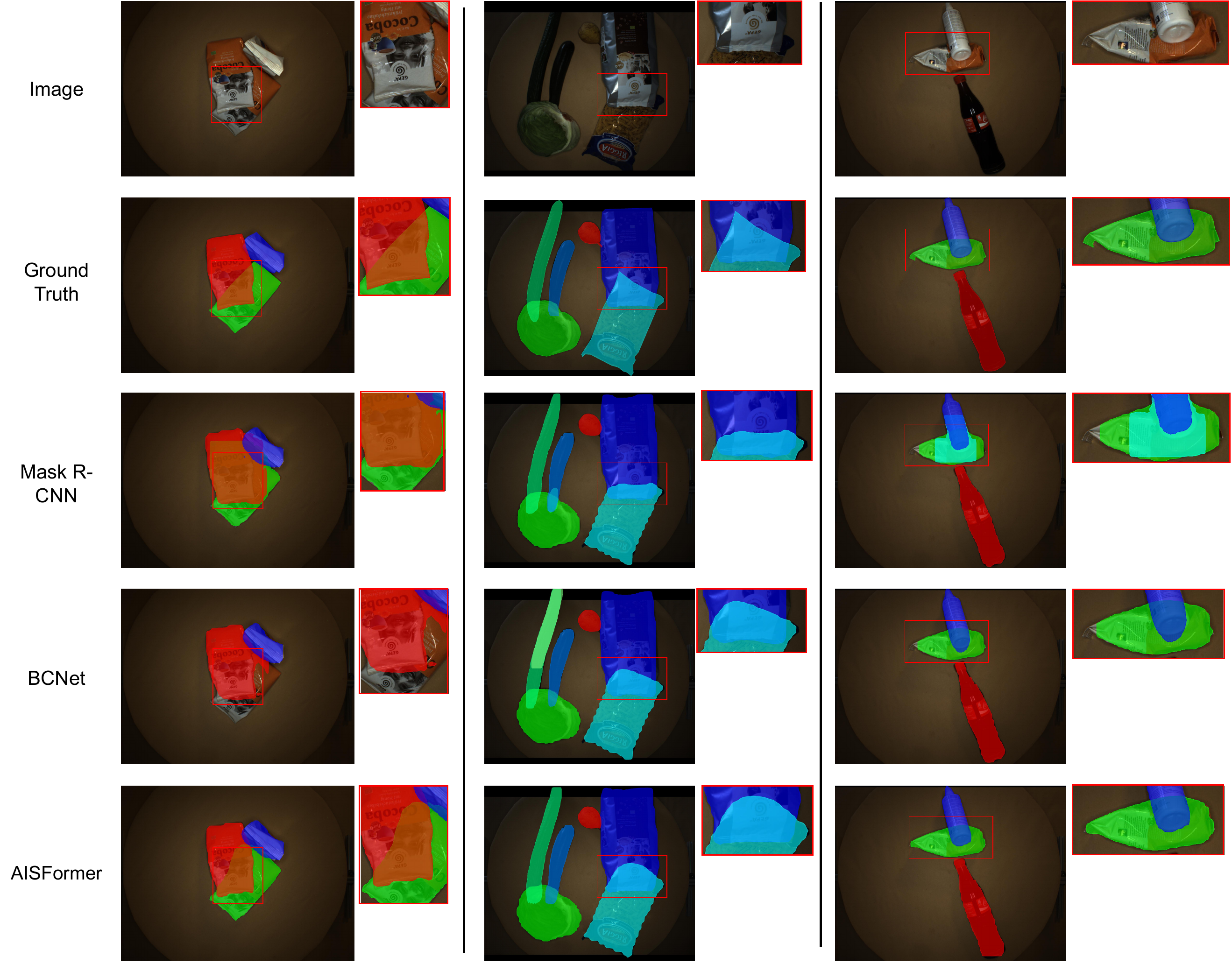}
    \vspace{0.2cm}
    \caption{Qualitative Results on D2SA dataset. The rows from top to bottom are the images, the ground-truth amodal masks, and estimations of Mask R-CNN, BCNet, and our AISFormer, respectively. Differences between methods are best shown in red box regions. Best view in zoom and color.} 
    \label{fig:qual_d2sa}
    \vspace{-0.3cm}
\end{figure*}

\begin{figure*}[t]
    \centering
    \includegraphics[width=\textwidth]{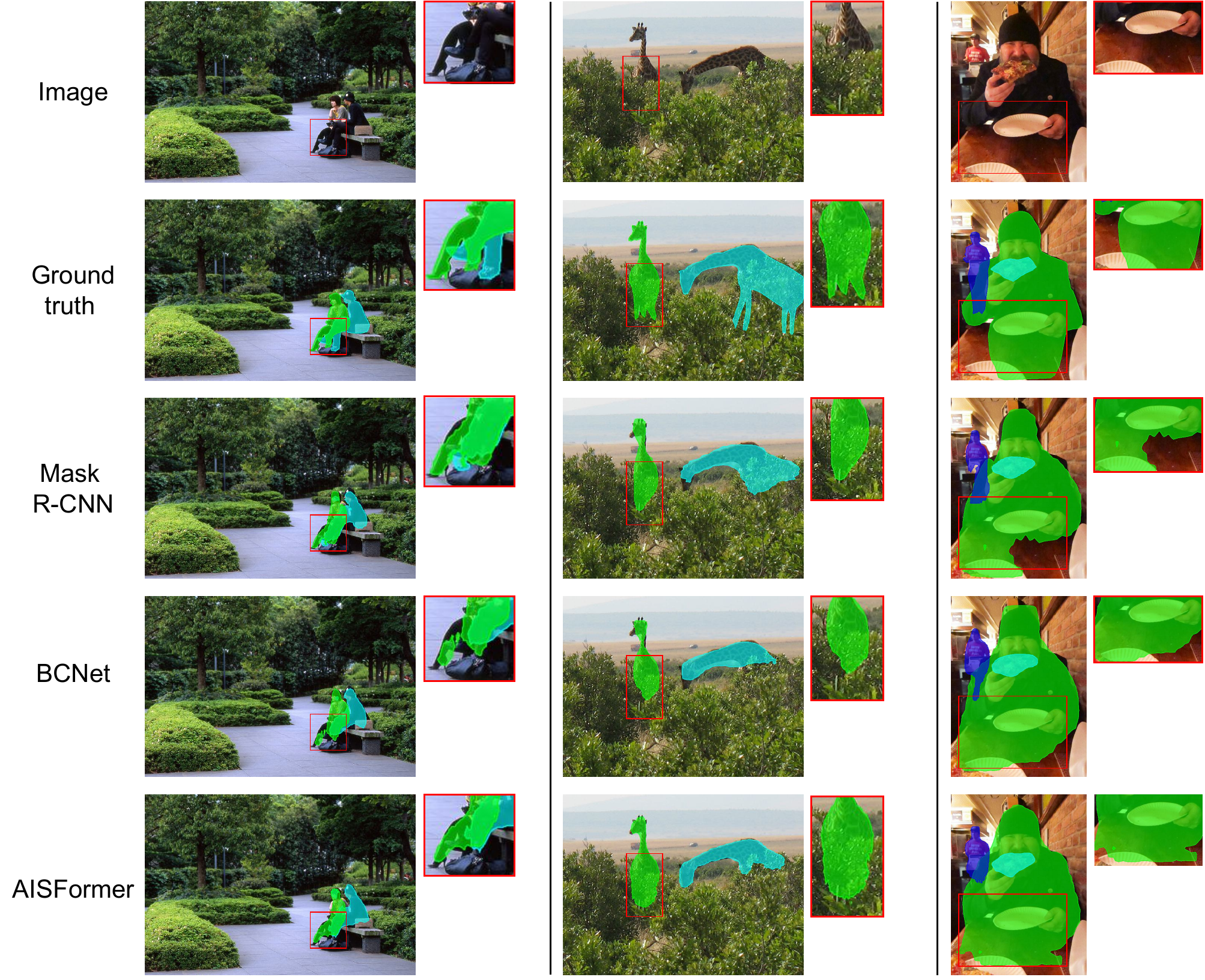}
    \vspace{0.2cm}
    \caption{Qualitative Results on COCOA-cls dataset. The rows from top to bottom are the images, the ground-truth amodal masks, and estimations of Mask R-CNN, BCNet, and our AISFormer, respectively. Differences between methods are best shown in red box regions. Best view in zoom and color.} 
    \label{fig:qual_cocoa}
    \vspace{-0.3cm}
\end{figure*}

Fig \ref{fig:qual_kins}, \ref{fig:qual_d2sa}, and \ref{fig:qual_cocoa} show our qualitative results compared with Mask R-CNN \cite{he2017mask} and BCNet \cite{ke2021deep} on the KINS\cite{qi2019amodal}, D2SA\cite{follmann2018mvtec} and COCOA-cls\cite{zhu2017semantic} datasets, respectively. As can be seen, in general, our AISFormer can segment more accurately than other methods, owning to the attention mechanism modeled between the amodal mask and other related masks (i.e. occluder mask and visible mask). 

In terms of Fig \ref{fig:qual_kins}, regarding the segmentation shown in the red box regions, for the first and second columns, our method is not affected by the occlusion and generates very fine masks of the back of the occluded cars. Moreover, in the last column in Fig \ref{fig:qual_kins}, our AISFormer maintains robustness in comparison with other methods despite the large occlusion rate. 

In terms of Fig. \ref{fig:qual_d2sa}, as illustrated in the first column, our method does a great job of distinguishing multiple objects that are overlapping each other. Meanwhile, in the last column, regarding the red box region, our AISFormer performs pretty accurate segmentation masks of the white bottle and the bag, which are better than those of Mask R-CNN, BCNet, and very close to the ground truth. 

In terms of Fig \ref{fig:qual_cocoa}, our AISFormer yields more accurate and reasonable segmentation masks in comparison with Mask R-CNN, BCNet, and ground truth. As can be seen in the first column image, our AISFormer predicts the more accurate mask of the occluded women, compared with the predicted masks of Mask R-CNN and BCNet. Moreover, in the last two columns, in comparison with Mask R-CNN and BCNet, our method yields more reasonable masks of the giraffe (second column) and the man who eats pizza (third column).

\subsection*{Attention visualizations}
\begin{figure*}[t]
    \centering
    \includegraphics[width=0.9\textwidth]{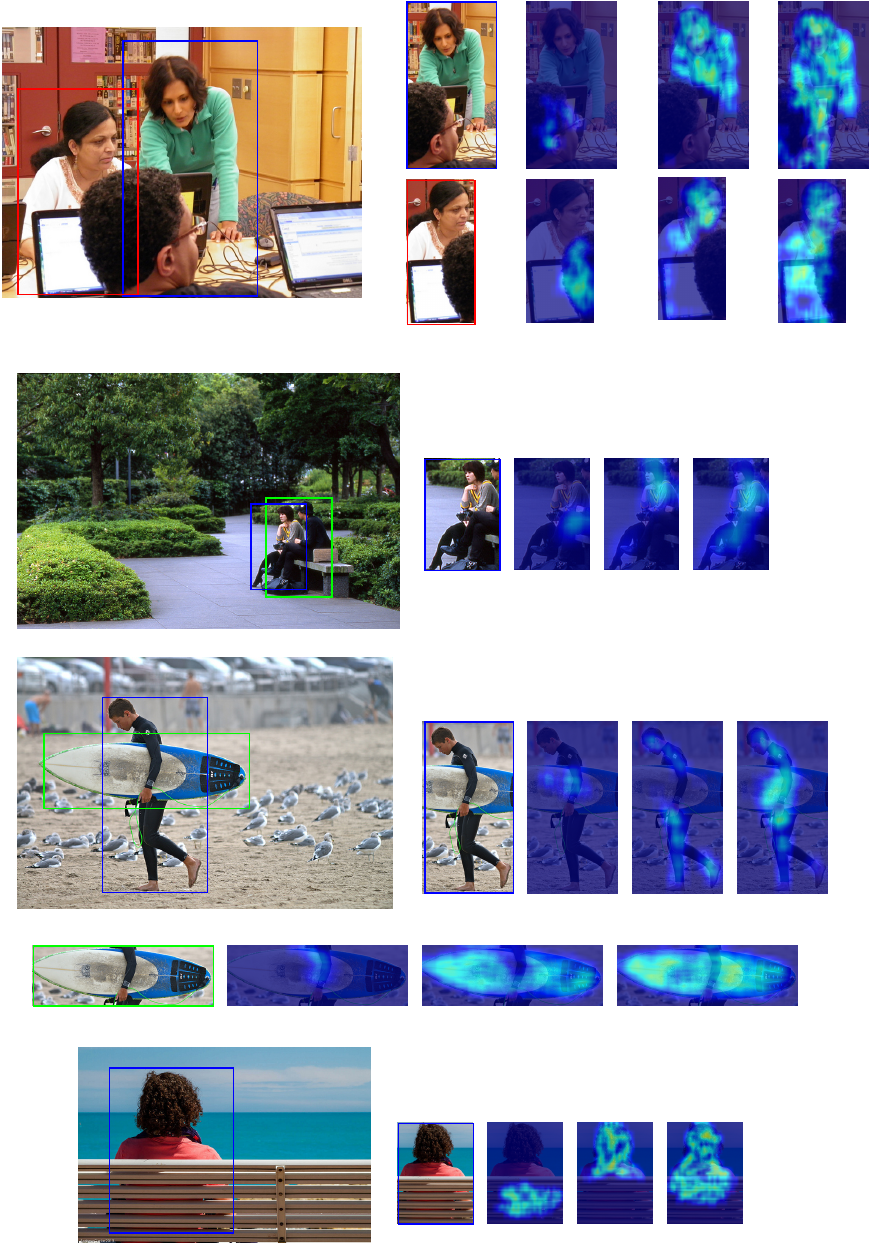}
    \vspace{0.2cm}
    \caption{Attention map visualizations indicating the cross attention between our learnable queries and the corresponding region of interest. For each ROI, from left-right: ROI, occluder query embedding, visible query embedding, and amodal query embedding} 
    \label{fig:attn_supp}
    \vspace{-0.3cm}
\end{figure*}

We provide more attention map visualization indicating the cross attention between our learnable queries and the corresponding region of interest in Fig \ref{fig:attn_supp}. 
